\documentclass{article}
\usepackage{ijcai19}

\usepackage{times}
\usepackage{soul}
\usepackage{url}
\usepackage[hidelinks]{hyperref}
\usepackage[utf8]{inputenc}
\usepackage[small]{caption}
\usepackage{graphicx}
\usepackage{amsmath}
\usepackage{booktabs}
\usepackage{amsthm}
\usepackage{booktabs}
\usepackage{algorithm}
\usepackage{algorithmic}
\usepackage[switch]{lineno}
\usepackage{amsfonts}

\newtheorem{definition}{Definition}

\title{Group-Aware Coordination Graph for Multi-Agent Reinforcement Learning}

\author{
Wei Duan\and
Jie Lu\and
Junyu Xuan\\
\affiliations
Australian Artificial Intelligence Institute (AAII), University of Technology\\
\emails
wei.duan@student.uts.edu.au,
\{jie.lu, junyu.xuan\}@uts.edu.au
}

\begin{document}

\maketitle

\begin{abstract}

Cooperative Multi-Agent Reinforcement Learning (MARL) necessitates seamless collaboration among agents, often represented by an underlying relation graph. Existing methods for learning this graph primarily focus on agent-pair relations, neglecting higher-order relationships. While several approaches attempt to extend cooperation modelling to encompass behaviour similarities within groups, they commonly fall short in concurrently learning the latent graph, thereby constraining the information exchange among partially observed agents. To overcome these limitations, we present a novel approach to infer the Group-Aware Coordination Graph (GACG), which is designed to capture both the cooperation between agent pairs based on current observations and group-level dependencies from behaviour patterns observed across trajectories. This graph is further used in graph convolution for information exchange between agents during decision-making.
To further ensure behavioural consistency among agents within the same group, we introduce a group distance loss, which promotes group cohesion and encourages specialization between groups. Our evaluations, conducted on StarCraft II micromanagement tasks, demonstrate GACG's superior performance. An ablation study further provides experimental evidence of the effectiveness of each component of our method.

\end{abstract}

\section{Introduction}

Cooperative Multi-Agent Reinforcement Learning (MARL) requires agents to coordinate with each other to achieve collective goals 
\cite{DBLP:journals/twc/CuiLN20,DBLP:journals/tits/0017W0H22}. To address the challenges of the expansive action space posed by multi-agents \cite{DBLP:journals/sensors/OrrD23}, a straightforward approach is breaking down the global training objective into manageable parts for each agent. Methods like  VDN \cite{DBLP:conf/atal/SunehagLGCZJLSL18}, QMIX \cite{DBLP:conf/icml/RashidSWFFW18} empower individual agents to select actions that maximize their own value function, contributing to the overall reward maximization. Nevertheless, it is important to recognize that many real-world tasks necessitate complex agent interactions and are not readily decomposable into simpler, individual tasks. This realization underscores the need to depict agent relationships, often assumed to involve latent graph structures in MARL. As the graph is not explicitly given, inferring dynamic graph topology remains a significant and persistent challenge.

\begin{figure}[t]
\centering
\includegraphics[width=0.9\columnwidth]{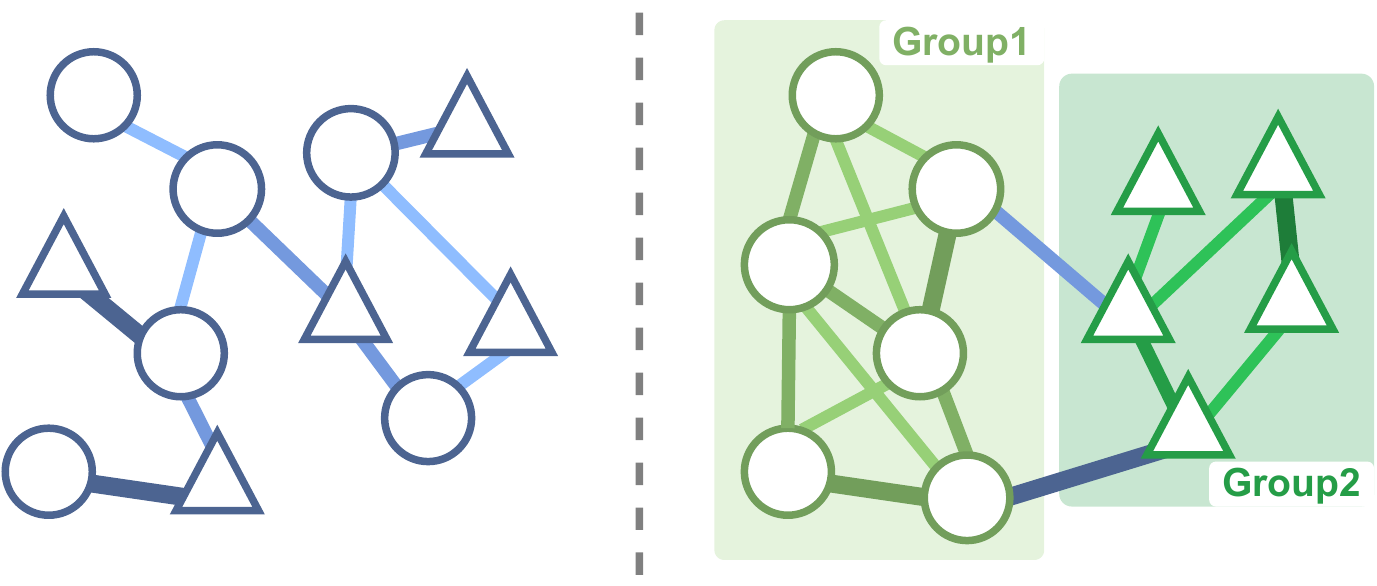} 
\caption{In a multi-agent environment, agents may exhibit diverse behaviours represented by triangles and circles. Existing methods for modelling agent interactions primarily focus on agent-pair relations. Concurrently recognizing the importance of higher-order group relationships among agents in coordination graphs is critical.}
\label{fig:motivation}
\end{figure}

\begin{figure*}[t]
\centering
\includegraphics[width=\textwidth]{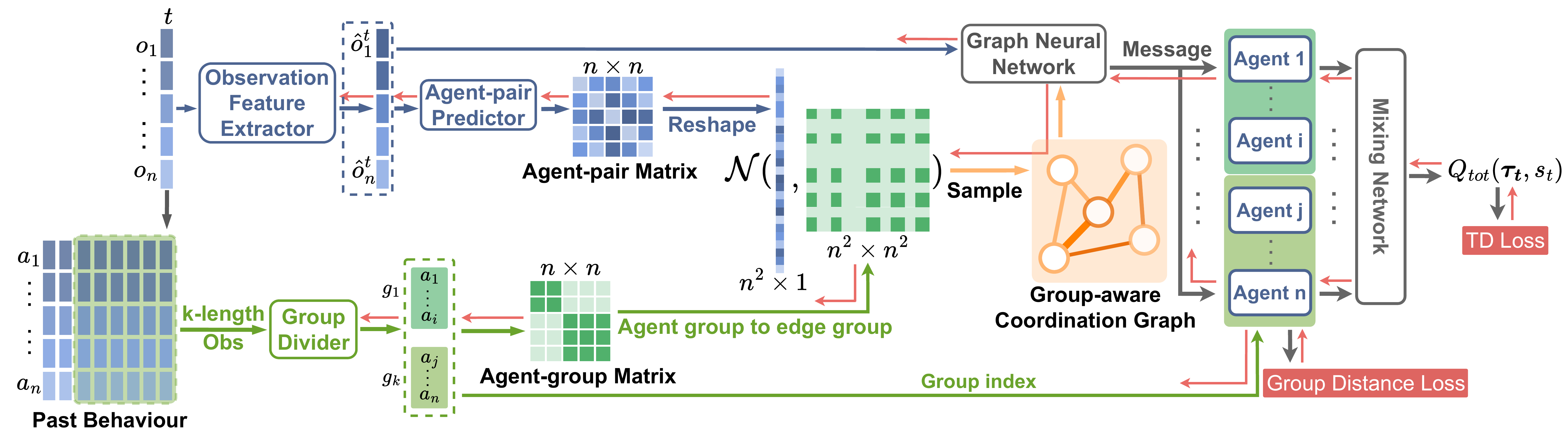} 
\caption{The framework of our method. GACG is designed to calculate cooperation needs between agent pairs based on current observations and to capture group-level dependencies from behaviour patterns observed across trajectories. All edges in the coordination graph as a Gaussian distribution. This graph helps knowledge exchange between agents when making decisions.  During agent training, the group distance loss regularizes behaviour among agents with similar observation trajectories.}
\label{fig:framework}
\end{figure*}

Current methods for learning the underlying graph in MARL can be categorized into three types: creating complete graphs by directly linking all nodes/agents \cite{liu2020pic,DBLP:conf/icml/BoehmerKW20}, employing attention mechanisms to calculate fully connected weighted graphs \cite{DBLP:conf/atal/LiGMAK21,DBLP:conf/aaai/LiuWHHC020}, and designing drop-edge criteria to generate sparse graphs \cite{DBLP:conf/icml/YangDRW0Z22,DBLP:conf/iclr/00010DY0Z22}.
However, these methods focus exclusively on agent-pair relations when modelling interactions. In multi-agent scenarios, such as orchestrating multi-robot formations \cite{DBLP:journals/csur/RizkAT19} or controlling a group of allies in strategic multi-agent combats \cite{samvelyan19smac}, relying solely on pairwise relations is inadequate for comprehensively understanding collaboration.  A critical aspect often overlooked is the importance of higher-order relationships, including group relationships/dependencies. Advancements have recently emerged that utilize group division, aiming to explore value factorization for sub-teams \cite{DBLP:conf/nips/PhanRBAGL21} or to specialize the character of the different group \cite{DBLP:conf/icml/IqbalWPBWS21}. Despite their efficacy in group partitioning, these methods do not concurrently learn the underlying graph structure. This limitation significantly affects the efficiency of information exchange among partially observed agents, which is vital for precise coordination and informed decision-making.

In light of these limitations, this paper proposes a novel approach to infer the Group-Aware Coordination Graph (GACG). GACG is designed to calculate cooperation needs between agent pairs based on current observations and to capture group dependencies from behaviour patterns observed across trajectories. A key aspect of our methodology is representing all edges in the coordination graph as a Gaussian distribution. This approach not only integrates agent-level interactions and group-level dependencies within a latent space but also transforms discrete agent connections into a continuous expression. Such a transformation is instrumental in modelling the uncertainty in various relationship levels, leading to a more informative and comprehensive representation of cooperation. Following this approach, the GACG is sampled from the distribution and used in graph convolution for information exchange between agents through a graph neural network during decision-making.  To further ensure behavioural consistency among agents within the same group, we introduce a group distance loss. This loss function is designed to increase the differences in behaviour between groups while minimizing them within groups. By doing so, it promotes group cohesion and encourages specialization between groups.

Experimental evaluations on StarCraft micromanagement tasks demonstrate GACG's superior performance. Our ablation study provides experimental evidence for the effectiveness of each component of our method.
The contributions of this paper are summarized as follows:
\begin{itemize} 
    \item  We propose a novel MARL approach named the Group-Aware Coordination Graph (GACG). To the best of our knowledge, this is the first method to simultaneously calculate cooperation needs between agent pairs and capture group-level dependencies within a coordination graph.
    \item  The edges of GACG are expressed as a Gaussian distribution, which models the uncertainty in various relationship levels, leading to a more informative and comprehensive representation of cooperation.
    \item We introduce the group distance loss to regularize behaviour among agents with similar observation trajectories, which enhances group cohesion and fosters distinct roles between different groups.
\end{itemize}

\section{Related Work}
The relationships among agents can be assumed to have latent graph structures \cite{DBLP:conf/iclr/TacchettiSMZKRG19,DBLP:conf/atal/LiGMAK21}. Graph Convolutional Networks (GNN) have demonstrated remarkable capability in modelling relational dependencies \cite{DBLP:conf/iske/DuanX021,DBLP:journals/tnn/WuPCLZY21,DBLP:conf/aaai/DuanXQ022,duan2024layerdiverse,DBLP:journals/eswa/YaoHLDQYS25}, making graphs a compelling tool for facilitating information exchange among agents \cite{DBLP:conf/iclr/0001WZZ20,DBLP:conf/aaai/LiuWHHC020} and serving as coordination graphs during policy training \cite{DBLP:conf/icml/BoehmerKW20,DBLP:conf/iclr/00010DY0Z22}. Although the current methods use different graph structures, such as a complete graph \cite{DBLP:conf/iclr/JiangDHL20,liu2020pic}, weighted graph \cite{DBLP:conf/icml/0001DLZ20} and sparse graph \cite{DBLP:conf/icml/YangDRW0Z22,DBLP:journals/corr/abs-2403-19253}, they primarily focus on agent-pair relations for inferring graph topology, neglecting higher-order relationships among agents.

Moving beyond the individual agent level, attention to group development becomes pivotal for maintaining diversified policies and fostering efficient collaborations. ROMA \cite{DBLP:conf/icml/0001DLZ20} learns dynamic roles that depend on the context each agent observes. VAST \cite{DBLP:conf/nips/PhanRBAGL21} approximates sub-group factorization for global reward to adapt to different situations. REFIL \cite{DBLP:conf/icml/IqbalWPBWS21} randomly group agents into related and unrelated groups, allowing exploration of specific entities. SOG \cite{DBLP:conf/nips/ShaoLZJHJ22} selects conductors to construct groups temporally, featuring conductor-follower consensus with constrained communication between followers and their respective conductors. GoMARL \cite{zang2023automatic} uses a “select-and-kick-out” scheme to learn automatic grouping without domain knowledge for efficient cooperation.  Despite their efficacy in group partitioning during training, these methods do not concurrently learn the underlying graph structure, thereby hindering the transfer of information within or between groups.

\section{Background}
We focus on cooperative multi-agent tasks modelled as a Decentralized Partially Observable Markov Decision Process (Dec-POMDP) \cite{DBLP:series/sbis/OliehoekA16} consisting of a tuple $\langle \mathcal{A}, \mathcal{S},\{\mathcal{U}_{i}\}_{i=1}^{n}, P,\{\mathcal{O}_{i}\}_{i=1}^{n},\{\pi_{i}\}_{i=1}^{n}, R,\gamma\rangle$, where $\mathcal{A}$ is the finite set of $n$ agents, $s \in \mathcal{S}$ is the true state of the environment. At each time step, each agent $a_i$ observes the state partially by drawing observation $o_{i}^{t} \in \mathcal{O}^{i}$ and selects an action $u_{i}^{t}\in \mathcal{U}^{i}$ according to its own policy $\pi_{i}$. Individual actions form a joint action $\boldsymbol{u}=(u_1,...,u_n)$, which leads to the next state $s^{\prime}$ according to the transition function $P(s^{\prime}| s, \boldsymbol{u})$ and a reward $R(s,\boldsymbol{u})$ shared by all agents. 
This paper considers episodic tasks yielding episodes $(s^0,\{o_{i}^{0}\}_{i=1}^{n},\boldsymbol{u}^0,r^{0},...,s^T,\{o_{i}^{T}\}_{i=1}^{n})$ of varying finite length $T$. Agents learn to collectively maximize the global return $Q_{tot}(s, \boldsymbol{u})=\mathbb{E}_{s_{0: T}, u_{0: T}}\left[\sum_{t=0}^{T} \gamma^t R\left(s^t, \boldsymbol{u}^t\right) \mid s^0=s, \boldsymbol{u}^0=\boldsymbol{u}\right]
$, where $\gamma \in [0,1)$ is the discount factor.

\section{Method}

The framework of our method is depicted in Fig. \ref{fig:framework}, which is designed to calculate cooperation needs between agent pairs based on current observations and to capture group-level dependencies from behaviour patterns observed across trajectories. This graph helps agents exchange knowledge when making decisions.  During agent training, the group distance loss regularizes behaviour among agents with similar observation trajectories, which enhances group cohesion and encourages specialization between groups.

\subsection{Group-Aware Coordination Graph Inference}

\begin{definition}
(Coordination graph (CG)). Given a cooperative task with $n$ agents, the coordination graph is defined as $ \mathcal{C}=\{\mathcal{A},\mathcal{E}\}$, where $\mathcal{A}=\left\{a_1, \ldots, a_n\right\}$ are agents/node and $\mathcal{E}=\left\{e_{11}, \ldots, e_{nn}\right\}$ edges/relations between agent. $|\mathcal{E}|= n^2$ indicates the number of possible edges. CG can be written in an adjacent matrix form as $C$.
\end{definition}

To effectively capture the evolving importance of interactions between agents, our approach leverages two components: the observation feature extractor $f_{oe}(\cdot)$ and agent-pair predictor $f_{ap}(\cdot)$. These components are designed to extract hidden features from the current observations of all agents at time $t$ and transform them into a meaningful structure we term the  \textbf{agent-pair matrix}:
\begin{equation}
\hat{o}^t_i = f_{oe}(o^{t}_{i}), \quad \boldsymbol{\mu}^{t}_{ij} = f_{ap}(\hat{o}^t_i,\hat{o}^t_j),
\label{eq:obs-emb}
\end{equation}
where $f_{oe}(\cdot)$ is realized as a multi-layer perceptron (MLP), $f_{ap}(\cdot)$ is an attention network. The dimension of agent-pair matrix $\boldsymbol{\mu}^{t}$ is $n \times n$, and it represents the edge weights for each agent pair, indicating the importance of their interaction at time $t$.

In a multi-agent environment, understanding the dynamics solely through pairwise relations is insufficient. To address this, we introduce the group concept, allowing us to extract higher-level information for more informed cooperative strategies among agents. We define a group as:

\begin{definition}
(Individual and Group). Given $n$ agents $\mathcal{A}$, we have a set of groups $\mathcal{G}=\left\{g_1, \ldots, g_m\right\}, 1 \leq m \leq n$. Each group $g_i$ contains $n_i\left(1 \leq n_i \leq n\right)$ different agents $g_i=\left\{a_1^i, \ldots, a_{n_i}^i\right\} \subseteq \mathcal{A}$, where $g_i \cap g_j=\emptyset , i \neq j, \cup_i g_i=\mathcal{A}$, and $i, j \in[1, m]$. 
\end{definition}

Given the inherent partial observability in MARL, we assume that agents with similar observations over specific time periods are likely to encounter similar situations, leading to similar behaviours. Following this premise, we introduce the group divider model represented by $f_g: \mathcal{A} \mapsto \mathcal{G}$, which aims to capture similarities among agents based on behaviour patterns observed across trajectories:
\begin{equation}
    \mathcal{G} = f_g (\mathcal{O}^{t-k:t}),
\end{equation}
where $\mathcal{O}^{t-k:t}$ denotes the observations of all agents from time $t-k$ to $t$. The parameter $k$ provides flexibility in determining the duration of the time steps, which allows us to choose the length of the trajectory. The reasons for using $\mathcal{O}^{t-k:t}$ to indicate trajectory behaviour are twofold: firstly, group dependencies are often observed over a time period, as in scenarios like coordinating a group of allies in an attack, where observations (such as facing the enemy) and behaviours (like attacking) tend to similar until the objective is achieved. Secondly, historical trajectory data has been shown to represent agents' behaviours more accurately than one-step observations \cite{pmlr-v119-pacchiano20a}, making it a more realistic and reliable source for our group divider.

Once groups are determined, we calculate the agent-group matrix $\mathbf{M}$ to indicate whether two agents belong to the same group at time $t$. This matrix is crucial for understanding group relations, which is defined as:
\begin{equation}
    \mathbf{M}^{t}_{ij} = \begin{cases} 
        1 & \text{if } a_i, a_j \in g_{m}  \text{ at time } t \\
        0 & \text{otherwise}.
    \end{cases}
\end{equation}

To effectively calculate cooperation needs between agent pairs based on current observations and capture group dependencies from behaviour patterns observed across trajectories, we model all edges as a Gaussian distribution. Specifically, the mean values of this distribution indicate the importance of interactions between agent pairs, where a larger mean value indicates a stronger interaction. The covariance matrix of this distribution encapsulates the dependencies between edges. This feature becomes particularly crucial when agents are part of the same group, as it underscores a heightened level of dependence among them.

Building on this idea, we first convert the agent-group matrix into the edge-group matrix, enabling direct incorporation of group information into edge relationships. This is described by the following operation:

\begin{equation}
    \mathbf{\hat{M}}^{t} =\text{vec}(\mathbf{M}^{t}) \times \text{vec}(\mathbf{M}^{t})^{\top}, \quad \boldsymbol{\Sigma} = \mathbf{\hat{M}},
\label{eq:edge-matrix}
\end{equation}
where the shape of $\text{vec}(\mathbf{M}^{t})$ is $ n^2 \times 1$ (the number of possible edges) and $\mathbf{\hat{M}}^{t}$ is $n^2 \times n^2$. An element in $\mathbf{\hat{M}}^{t}$ being $1$ means that the corresponding edges belong to the same group.

\begin{definition}
(Edegs in the same group). Given two edge $e_{ij},e_{lk} \in\mathcal{E} $, if $ a_i, a_j, a_l, a_k \in g_m$, we assert that $ e_{ij}$ and $ e_{lk}$ in same group.
\end{definition}

Utilizing the \textbf{agent-pair matrix} $\boldsymbol{\mu}^{t}$ and \textbf{edge-group matrix }$\mathbf{\hat{M}}^{t}$, the Gaussian distribution is formally represented as:
\begin{equation}
    \mathcal{E} \sim \mathcal{N}(\boldsymbol{\mu}^{t}, \mathbf{\hat{M}}^{t}),
\label{eq:G-dis}
\end{equation}
where the shapes of $\boldsymbol{\mu}^{t}$ and
$\mathbf{\hat{M}}^{t}$ are $n^2 \times 1$ and $n^2 \times n^2$, respectively. This approach not only provides a practical tool for capturing edge dependencies within the graph but also modelling the uncertainty in various agent relationship levels. For instance, considering two edges $e_i, e_j \in \mathcal{E}$, they follow the distribution (one property of multivariate Gaussian distribution):
\begin{equation}
    (e_i, e_j) \sim \mathcal{N}\left((\mu^{t}_{i}, \mu^{t}_{j}), \left[\begin{array}{ll}
\mathbf{\hat{M}}^{t}_{ii} & \mathbf{\hat{M}}^{t}_{ij} \\
\mathbf{\hat{M}}^{t}_{ji} & \mathbf{\hat{M}}^{t}_{jj}
\end{array}\right]\right).
\label{eq:G-dis-edges}
\end{equation}
According to the above definitions, if  $e_i$ and $e_j$ are in the same group (as indicated by $\mathbf{\hat{M}}^{t}_{ij}=1$), they will exhibit high dependence, implying a closely aligned probability of their simultaneous occurrence or absence. Conversely, if the edges belong to different groups ($\mathbf{\hat{M}}^{t}_{ij}=0$), their existences are modelled as independent. This approach aligns with the expectation that edges within the same group should exhibit stronger contextual dependencies, enhancing the model's capability to accurately represent and adapt to complex interactions among agents.

\begin{figure*}[t]
\centering
\includegraphics[width=\textwidth]{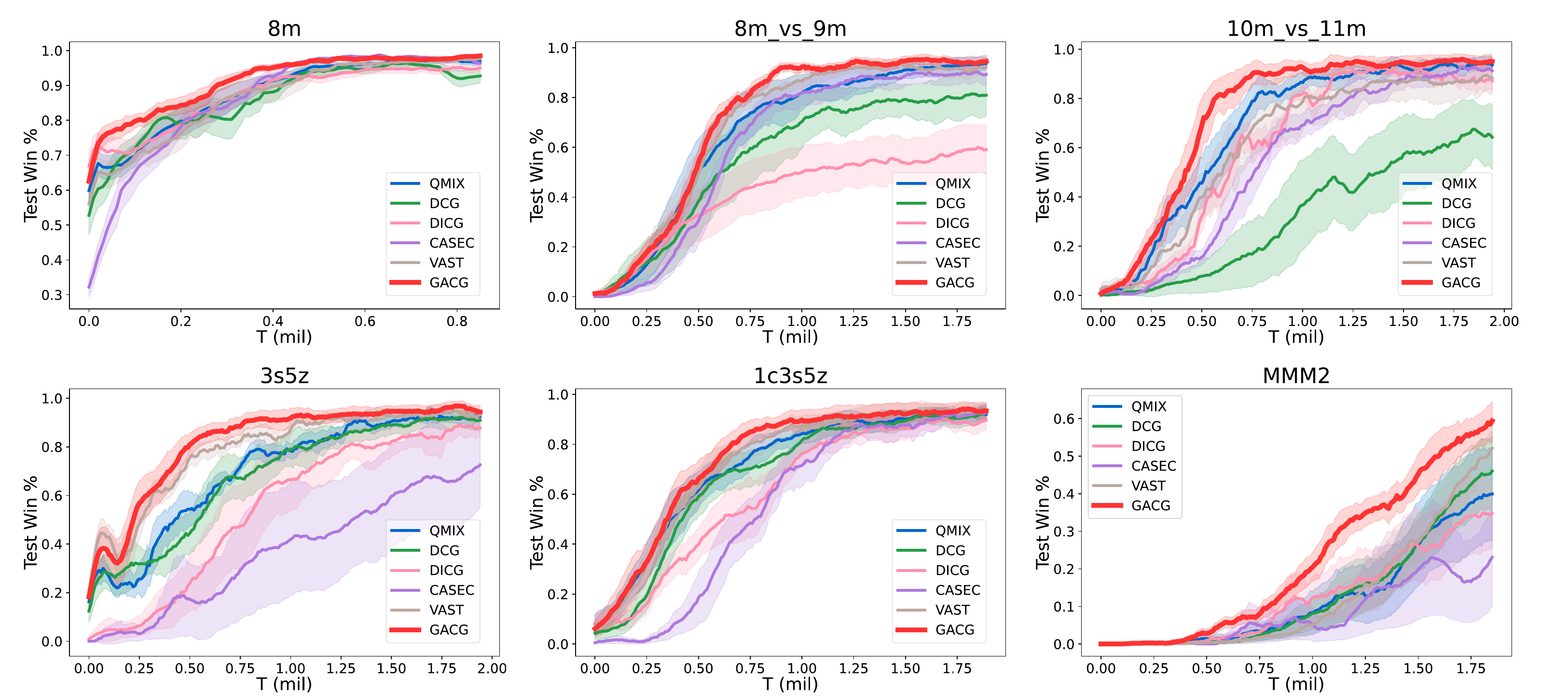} 
\caption{Performance of GACG and baselines on six maps of the SMAC. The x-axis represents the time steps (in millions), while the y-axis quantifies the test win rate in the games..}
\label{fig:Result}
\end{figure*}

\subsection{Group-Aware Cooperative MARL}
Utilizing the Gaussian distribution defined in Eq.(\ref{eq:G-dis}), we sample the edges of the Group-Aware Coordination Graph at each timestep, reshaping them into matrix form $C^{t}$. This sampled graph structure facilitates information exchange between agents through a graph neural network (GNN) \cite{10255371}. The GNN's message-passing mechanism is essential for agents to efficiently share and integrate information, adapting their strategies to the dynamic MARL environment. The GNN is defined as follows:
\begin{equation}
\label{eq:GCN_MATIX}
   H^{t}_{l} = ReLU \Big( \hat{C}^{t} H^{t}_{(l-1)} W_{(l-1)} \Big), 
\end{equation}
where $l$ is the index of GNN layers, $\hat{C}^{t} = \tilde{D}^{-\frac{1}{2}}C^{t}\tilde{D}^{-\frac{1}{2}}$, $\tilde{D}_{ii} = \sum_{j} C^{t}[i,j]$. The initial input of the GNN, $H_{t}^{0}$, is set to the extracted observation features $\{\hat{o}^t_1,...,\hat{o}^t_n\}$ from Eq.(\ref{eq:obs-emb}). The output of GNN $m^{t}_i = H^{t}_{l} $ treated as exchanged knowledge between agents, which is then utilized in the local action-value function defined as $Q_i(\tau_{i},\mu_j,m^{t}_i)$. 

Building upon our earlier assumption that agents with similar trajectories are likely to exhibit similar behaviours, it becomes straightforward to regularize the behavioural consistency of agents during the policy training phase. This behaviour is reflected in the output of $\pi_i(\mu_i|\tau_i)$ (or $Q_i$),  representing the probability distribution of actions for the current state. To assess the similarity of behaviour among agents, we compare their policy outputs and introduce the group distance loss, defined as: 
\begin{equation}
    \mathcal{L}_{g}=\frac{\frac{1}{(m-1)^2} \sum_{i \neq j}\left(\frac{1}{\left|g_i\right|\left|g_j\right|} \sum_{ a_l \in g_i} \sum_{ a_k \in g_j}\left\|\pi_{l}-\pi_{k}\right\|_2\right)}{\frac{1}{m} \sum_i\left(\frac{1}{\left|g_i\right|^2} \sum_{a_{l}, a_{v} \in g_i}\left\|\pi_{l}-\pi_{v}\right\|_2\right)}.
\end{equation}
This equation calculates the average pairwise behavioural distances between agents of different groups (numerator) and within the same group (denominator). By minimizing intra-group distances, this loss function promotes uniform behaviour within groups while maximizing inter-group distances to encourage diversity and specialization.

Our algorithm is built on top of QMIX \cite{DBLP:conf/icml/RashidSWFFW18}, integrating all individual Q values for overall reward maximization.  The training involves minimizing a loss function, composed of a temporal-difference (TD) loss and the group distance loss, as follows:
\begin{equation}
 \mathcal{L}(\boldsymbol{\theta})=\mathcal{L}_{TD}(\boldsymbol{\theta}^{-})+\lambda \mathcal{L}_{g}\left(\boldsymbol{\theta}_g\right), 
\label{eq:final-loss}
\end{equation}
where  $\boldsymbol{\theta}$ includes all parameters in the model, $\lambda$ is the weight of group distance loss. The TD loss $\mathcal{L}_{TD}(\boldsymbol{\theta}^{-})$ is defined as 
\begin{equation}
\mathcal{L}_{TD}(\boldsymbol{\theta}^{-})\!=\!\left[r\!+\!\gamma \max _{\boldsymbol{a}^{\prime}} Q_{tot}\left(s^{\prime}, \boldsymbol{\mu}^{\prime} ; \boldsymbol{\theta}^{\prime}\right)\!-\!Q_{tot}(s, \boldsymbol{\mu} ; \boldsymbol{\theta}^{-})\right]^2,
\label{eq:loss}
\end{equation}
where $\boldsymbol{\theta}^{\prime}$ denotes the parameters of a periodically updated target network, as commonly employed in DQN. 


\section{Experiments}
In this section, we design experiments to answer the following questions: (1) How well does GACG perform on complex cooperative multi-agent tasks compared with other state-of-the-art CG-based methods? (2) Is the choice and calculation method for the Gaussian distribution promising for sampling edges in CG? (3) Does the inclusion of the group loss improve GACG's performance? (4) What is the influence of group number on the GACG performance? (5) How does the selected length of trajectory affect the final result?

The experiments in this study are conducted using the StarCraft II benchmark \cite{DBLP:conf/atal/SamvelyanRWFNRH19}, which offers complex micromanagement tasks with varying maps and different numbers of agents. The benchmark includes scenarios with a minimum of eight agents, encompassing both homogeneous and heterogeneous agent setups. The environments are configured with a difficulty level of 7, providing a rigorous testing ground for evaluating the performance and generalization capabilities of MARL algorithms. The experiments are systematically carried out with 5 random seeds to ensure robustness and reliability in the assessment of the proposed methods.

\begin{table}
    \centering
    \begin{tabular}{cccc}
        \toprule
        Method  & Graph type  & Edge  & Group \\
        \midrule
        QMIX     & $\times$  & $\times$   & $\times$\\
        DCG     & Complete & Unweighted   & $\times$ \\
        DICG & Complete & Weighted    & $\times$ \\
        CASEC & Sparse  & Weighted    & $\times$ \\
        VAST & $\times$  & $\times$   & $\surd$ \\
        \textbf{GACG} & \textbf{Sparse}  & \textbf{Weighted}     & $\boldsymbol{\surd}$ \\
        \bottomrule
    \end{tabular}
    \caption{Comparison of different experiment methods in terms of graph type, edge representation, and group utilization. }
    \label{tab:method}
\end{table}
\subsection{Compared with other CG-based Methods}

We compare our methods with the following baselines, and each method's graph type, edge representation, and group utilization are summarised in Tab.\ref{tab:method}.  
\begin{itemize}
    \item \textbf{QMIX}\footnote{https://github.com/oxwhirl/pymarl} \cite{DBLP:conf/icml/RashidSWFFW18} is effective but without cooperation between agents, also without group division.
    \item \textbf{DCG} \footnote{https://github.com/wendelinboehmer/dcg}\cite{DBLP:conf/icml/BoehmerKW20} directly links all the edges to get an unweighted fully connected graph. The graph is used to calculate the action-pair values function. 
    \item \textbf{DICG}\footnote{https://github.com/sisl/DICG} \cite{DBLP:conf/atal/LiGMAK21} uses attention mechanisms to calculate weighted fully connected graph. The graph is used for information passing between agents. 
    \item \textbf{CASEC}\footnote{https://github.com/TonghanWang/CASEC-MACO-benchmark} \cite{DBLP:conf/iclr/00010DY0Z22} drop edges on the weighted fully connected graph using the variance payoff function. 
    \item \textbf{VAST} \footnote{https://github.com/thomyphan/scalable-marl}\cite{DBLP:conf/nips/PhanRBAGL21}  explores value factorization for sub-teams based on a predetermined group number. The sub-team values are linearly decomposed for all sub-team members.
    
\end{itemize}

\subsubsection{Results}
In Figure \ref{fig:Result}, we present the comprehensive results of our experiments conducted across six diverse maps, highlighting the superior performance of our Group-Aware Coordination Graph (GACG) method. GACG consistently achieves high win rates with rapid convergence and reliability, outperforming competing methods in complex multi-agent scenarios.  On the \textit{8m} and \textit{1c3s5z} maps, all methods demonstrate similar convergence patterns. They reach their highest test win rates at the end of 2 million training steps. However, performance disparities become more evident on other maps. For instance, DICG exhibits the weakest performance on the \textit{8m\_vs\_9m} map, struggling to match the effectiveness of other methods. Similarly, DCG falls behind on the \textit{10m\_vs\_11m} map, where the competing approaches outperform it. On the challenging 3s5z map, CASEC shows a lower win rate, suggesting that its strategy is less suited to the intricacies of this particular scenario. 

The limitations of comparative methods are apparent: QMIX lacks graph structures, DCG treats all interactions uniformly without weight considerations, DICG misses group dynamics, and CASEC, despite addressing message redundancy, overlooks the importance of group-level behaviour. VAST, while exploring sub-team dynamics, does not utilize dynamic graph structures. GACG's nuanced approach—leveraging agent pair cooperation and group-level dependencies—affords a deeper understanding of agent interactions. The group distance loss embedded in training sharpens within-group behaviour and enhances inter-group strategic diversity, contributing to GACG's effectiveness.


In summary, the comparative analysis validates that a sophisticated approach to graph learning, attentive to pairwise and group-level information, is instrumental in achieving superior performance in MARL. With its innovative graph sampling technique and the seamless incorporation of group dynamics, the Group-Aware Coordination Graph method offers more intelligent and adaptable cooperative learning algorithms.

\subsubsection{Computational Complexity Analysis}
The computational complexity of our model is primarily influenced by the necessity to discern group relationships among agents. Given that interactions are represented as edges in a fully connected graph, an environment with $n$ agents results in $n^2$ pairwise edges. To capture group dynamics, an edge-group matrix $\mathbf{\hat{M}}^{t}$ is constructed to represent these relationships, with dimensions $n^2 \times n^2$. Consequently, the time complexity for computing the group matrix is $O(n^4)$.

While the theoretical computational complexity for computing the group matrix in our model seems high, this computation is mainly attributed to the multiplication operation $\mathbf{\hat{M}}^{t} =\text{vec}(\mathbf{M}^{t}) \times \text{vec}(\mathbf{M}^{t})^{\top}$ in Eq.(\ref{eq:edge-matrix}). Importantly, this operation is highly amenable to parallelization, a task at which GPUs excel, substantially accelerating the actual training process. Tab.\ref{tab:time} presents empirical running times for our model on the \textit{10m\_vs\_11m} map, demonstrating that despite the theoretically high complexity, our GACG method is competitive in practice. In fact, the running times for GACG are faster or on par with other graph-based methods.

\begin{table}[!t]
\begin{center}
\begin{tabular}{ccc}
\toprule
& $1k$ steps time $(s)$ & $1m$ steps time $(h)$  \\
\midrule
QMIX  & $20.13\pm3.59$     &   $6.79\pm0.37 $   \\
DCG   & $33.57\pm4.65$    & $11.63\pm0.64$     \\
CASEC &  $30.50\pm2.03$    &  $10.12\pm0.51$    \\
VAST & $21.28\pm3.59$     &   $6.21\pm0.56 $   \\
GACG &  $22.33\pm4.22$     &  $7.84\pm0.49 $     \\
\bottomrule
\end{tabular}
\end{center}
\caption{Time computational consumption on map \textit{10m\_vs\_11m}.}
\label{tab:time}
\end{table}

\subsection{Abaltion study}
\subsubsection{Different Edge distributions}
In this part, we experiment on two maps aiming to provide insights into the unique aspects of the Gaussian distribution in GACG and its effectiveness in capturing agent-level and group-level information for cooperative multi-agent tasks. We substitute this part with other distributions. The detail settings are shown below:

\begin{itemize}
\item \textbf{Attention (without distribution):} This approach uses no specific distribution for the coordination graph but learns the edges directly from the agent-pair matrix, where edges are obtained as  $e^{t}_{ij}=f_{ap}(\hat{o}^t_i,\hat{o}^t_j)$. 

\item \textbf{Bernoulli:} The distribution is changed to a Bernoulli distribution, where the agent-pair matrix serves as the probability of this distribution, expressed as $P_B(e_{ij}=1) = f_{ap}(\hat{o}^t_i,\hat{o}^t_j)$.

\item \textbf{Inde-Gaussian:} Each edge is considered independent and follows a Gaussian distribution. Each element in the agent-pair matrix serves as mean values for corresponding edges, formulated as $e_{ij} \sim \mathcal{N}(\boldsymbol{\mu}_{ij}, \sigma^2)$, where $0 \leq \sigma^2 \leq 1$. This setting investigates the impact of group-level dependence when designing the multivariate  Gaussian distribution.

\item \textbf{GACG (w/o $\mathcal{L}_g$):} The final loss function is without the inclusion of the group distance loss $\mathcal{L}_g$. This ensures that the only difference between the compared methods is distribution.
\end{itemize}





\begin{figure}[h]
\centering
\includegraphics[width=\columnwidth]{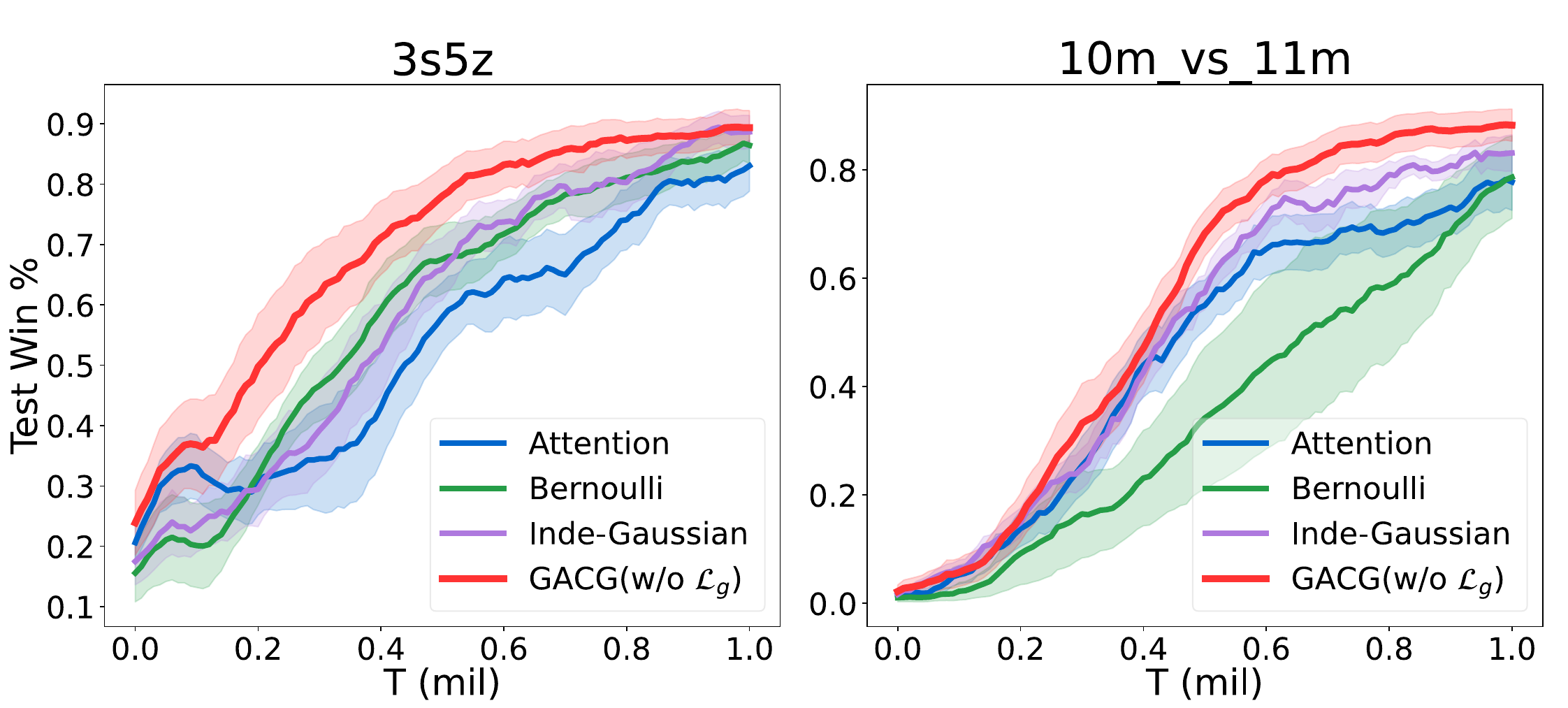} 
\caption{Experiment of choosing different edge distributions when learning the CG.}
\label{fig:Dsitribution}
\end{figure}

The result is shown in Fig.\ref{fig:Dsitribution}. Across various maps, our GACG consistently outperforms other settings, confirming the efficacy of our method. 
In the \textit{3s5z} map, regardless of the distribution used, treating the learning of the graph as the learning of the edge distribution is more effective than utilizing attention alone (without distribution). However, in the \textit{10m\_vs\_11m} map, the use of the Bernoulli distribution performs worse than attention, indicating that the choice of distribution is not arbitrary. This observation underscores the importance of carefully selecting the distribution method in constructing the coordination graph.

When compared with the setting where each edge is considered independent and follows a Gaussian distribution, our method yields better results. This finding emphasizes the importance and effectiveness of capturing group dependency when learning the edge distribution. The ability to model dependencies among agents at the group level contributes significantly to the improved performance of our approach.

\subsubsection{Effectiveness of Group Distance Loss.}
In this part, we test the effectiveness of group distance loss by training the GACG with and without $\mathcal{L}_g$.

\begin{figure}[t]
\centering
\includegraphics[width=\columnwidth]{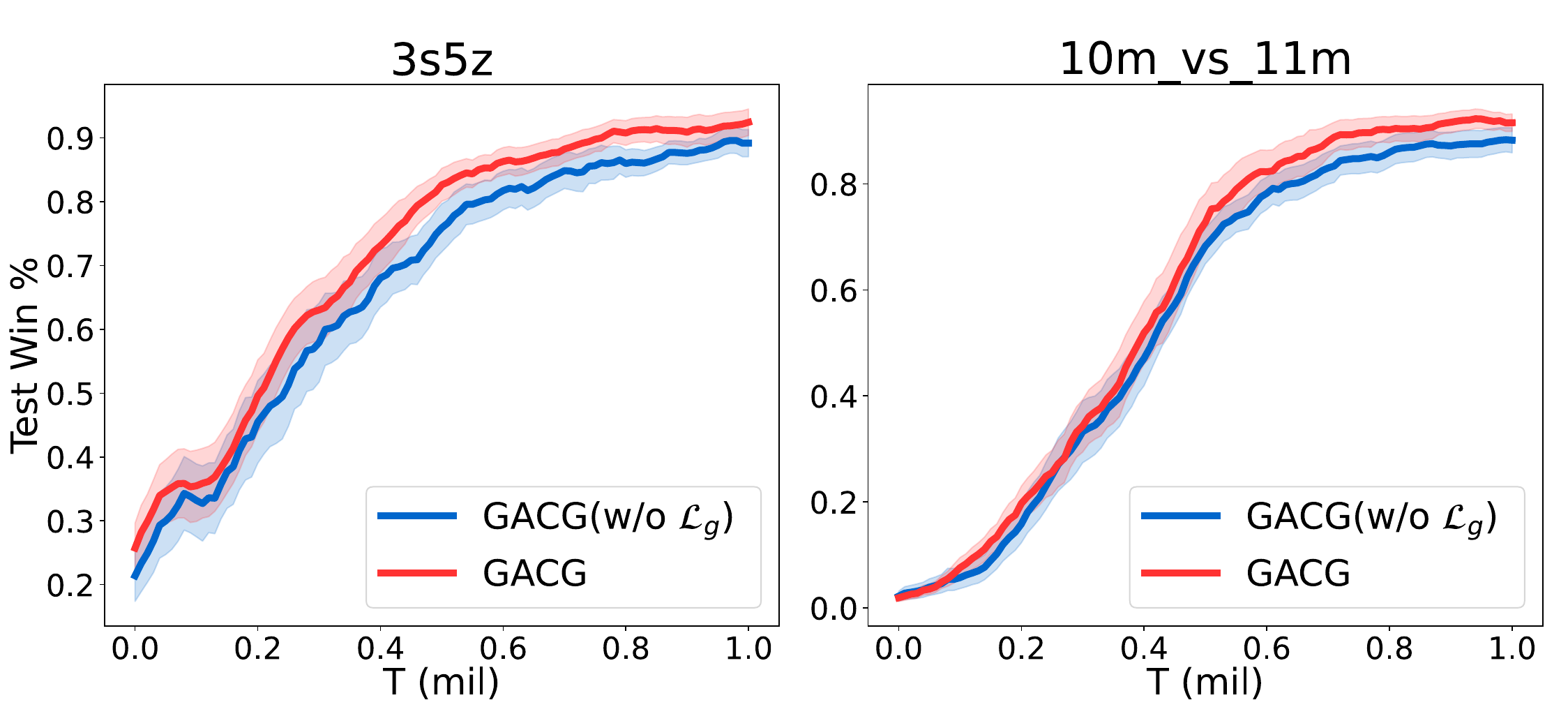} 
\caption{Experiment of training GACG with/without $\mathcal{L}_g$.}
\label{fig:GLoss}
\end{figure}

The results are shown in Fig.\ref{fig:GLoss}, revealing several key findings: (1) GACG trained with $\mathcal{L}_g$ exhibits a faster convergence speed and achieves a higher average final performance compared to its counterpart trained without $\mathcal{L}_g$. This observation strongly suggests that $\mathcal{L}_g$ effectively guides the model towards more efficient and cooperative learning. (2) This component contributes to enlarging inter-group distances while concurrently decreasing intra-group distances, fostering effective agent cooperation. (3) This result highlights the essential role that group-level information plays in enhancing the overall effectiveness and cooperation capabilities of GACG, demonstrating the significance of considering both agent-level and group-level dynamics for optimal performance.

\subsubsection{Number of Groups}
We further investigate the impact of varying the number of groups ($m$)  on two distinct maps: \textit{3s5z} and \textit{8m\_vs\_9m}, each featuring 8 agents. The former has two types of agents, while the latter consists of a single type. The experiment is conducted with $m$ set to $\{0,2,4,8\}$.

When $m=0$, it implies the absence of group divisions between agents, indicating that no group-level information is utilized during training (including graph reasoning and group loss calculation). Conversely, when $m=8$, each agent is treated as an individual group. In this setting, while there are no group divisions, the presence of the group distance loss encourages each agent to exhibit diverse behaviours and dissimilarity.

\begin{figure}[t]
\centering
\includegraphics[width=\columnwidth]{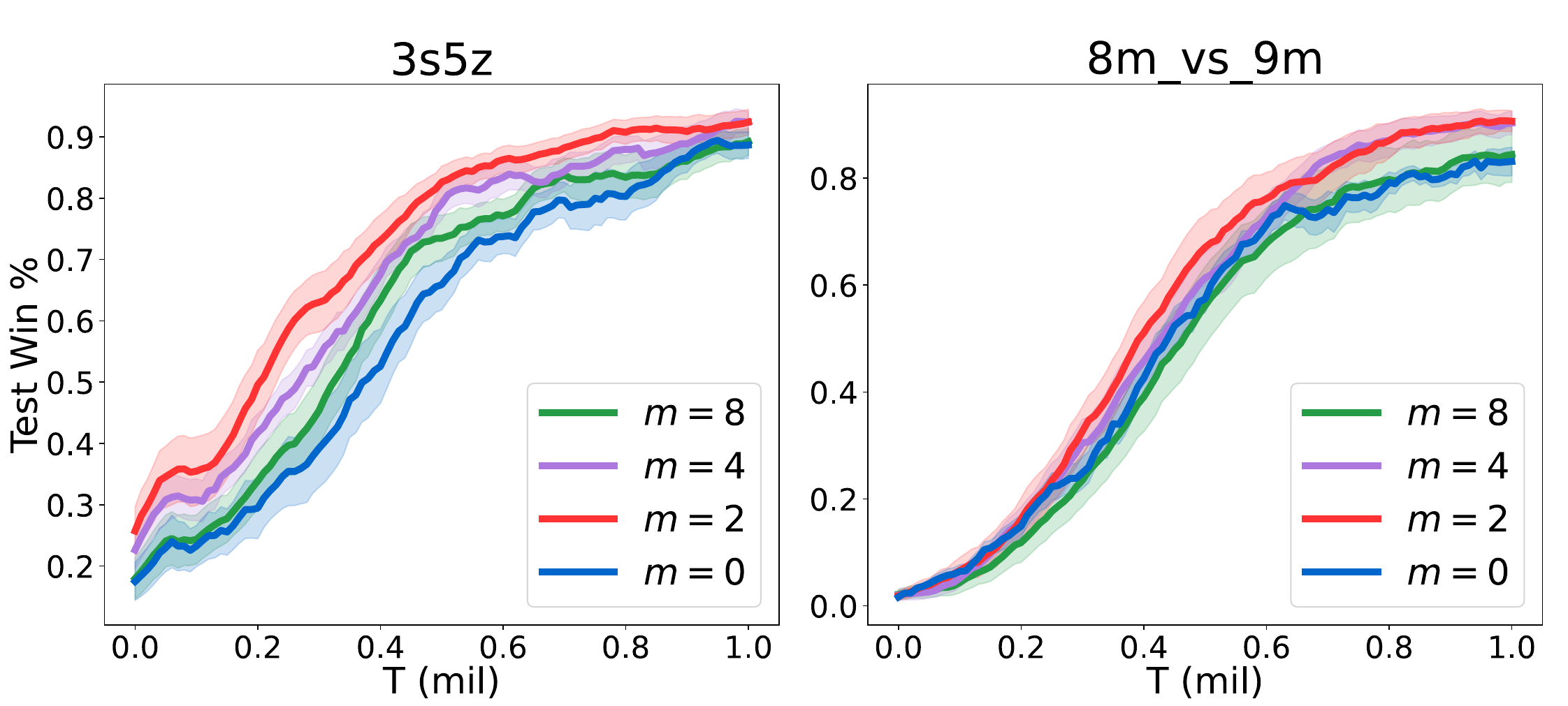} 
\caption{Experiment of dividing $n$ agents into different numbers of groups($m$) on \textit{3s5z} and \textit{8m\_vs\_9m}. The former has two types of agents, while the latter consists of a single type.}
\label{fig:Gnum}
\end{figure}

The results are illustrated in Fig. \ref{fig:Gnum}. Optimal performance is observed when $m=2$, underscoring the beneficial impact of a moderate level of group division on cooperation. Across both maps, the introduction of group divisions ($m \in {2, 4}$) consistently outperforms scenarios where there are either no group divisions or a high number of them ($m \in {0, 8}$). This emphasizes the crucial role of incorporating group-level information in achieving superior MARL outcomes. Notably, in map \textit{3s5z}, where two agent types are present, the convergence speed of $m=8$ is faster than that of $m=0$. This acceleration is likely due to the importance of policy diversity in a multi-agent setting with distinct agent types, and the group distance loss facilitates each agent in achieving diverse behaviours.

\subsubsection{Length of Trajectory for Group Division}
In this analysis, our objective is to investigate the influence of varying observation trajectory lengths, parameterized by $k$, on the effectiveness of the group divider model $f_g$. We explore different values for $k$, specifically $\{1, 5, 10, 20\}$. When $k=1$, a single timestep is considered, whereas $k \in \{5, 10, 20\}$ enables the assessment of longer temporal relationships. 

\begin{figure}[t]
\centering
\includegraphics[width=\columnwidth]{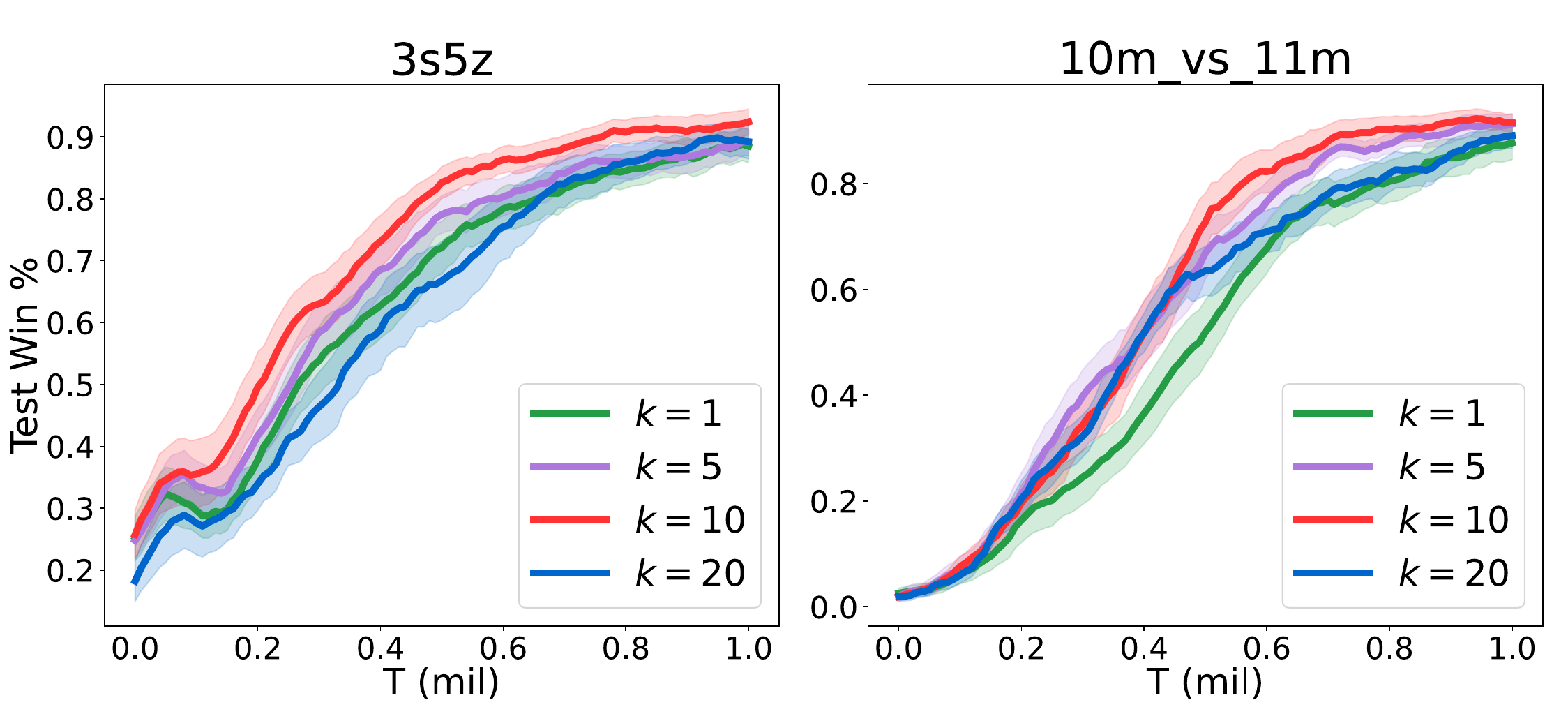} 
\caption{Experiment of varying observation window lengths ($k$) on the group divider model $f_g$.}
\label{fig:Olen}
\end{figure}

The results are depicted in Fig. \ref{fig:Olen}. Across both maps, the performance data indicates that a moderate observation window length of $k=10$ yields the highest test win percentage. This setting strikes a balance, providing agents with sufficient historical data to discern meaningful patterns and group relationships. Conversely, the $k=1$ setting, representing the shortest observation window, fails to deliver adequate historical data for effective group differentiation and decision-making, resulting in a lower test win percentage. As we increase the window length from $k=1$ to $k=5$, more information becomes available for group division, yet it does not surpass the performance achieved with $k=10$. 

An interesting observation arises when $k=20$, the overall performance decreases and approaches that of $k=1$. This phenomenon suggests a point of diminishing returns, where additional historical information may not contribute to better decision-making. This could be attributed to challenges such as overfitting or an inability to adapt quickly to new information. Therefore, selecting an optimal observation window, such as $k=10$, allows agents to integrate just enough temporal information. This enables adaptation to dynamic environments without the drawbacks associated with processing excessive or potentially noisy data. 

\section{Discussion}

Recent work has explored complementary aspects of structure and communication in cooperative MARL. Duan et al.\ \cite{duan2025bayesian} formulate Bayesian ego-graph inference for networked settings, focusing on inferring each agent's local interaction structure from limited observations. In contrast, GACG learns a global coordination graph together with trajectory-based group assignments, explicitly coupling pairwise edges with group-level regularities for graph convolution and policy learning. Duan et al.\ \cite{duan2026bandwidth} study bandwidth-constrained variational message encoding, which prioritizes compact, information-rich signals when agents communicate. GACG instead emphasizes \emph{who} should exchange information via the learned graph and group structure; combining such selective routing with bandwidth-aware encodings is a natural direction for scaling coordination to larger teams or communication-limited deployments.

\section{Conclusion}

In this paper, we have presented the Group-Aware Coordination Graph (GACG), a novel MARL framework that addresses the limitations of existing approaches. Unlike previous methods that handle agent-pair relations and dynamic group division separately, GACG seamlessly unifies the integration of pairwise interactions and group-level dependencies. It adeptly computes cooperation needs from one-step observations while capturing group behaviours across trajectories. Employing the graph's structure for information exchange during agent decision-making significantly enhances collaborative strategies. Incorporating a group distance loss during training enhances behavioural similarity within groups and encourages specialization across groups. Our extensive experimental evaluations reveal that our method consistently outperforms current leading methods. An ablation study confirms the efficacy of each individual component, highlighting the importance of incorporating both pairwise and group-level insights into the learning model. The outcomes of this research emphasize the importance of multi-level agent information integration, establishing our framework as a substantial contribution to advancing MARL.

\section*{Acknowledgements}
 This work is supported by the Australian Research Council under Australian Laureate Fellowships FL190100149 and Discovery Early Career Researcher Award DE200100245.

\bibliographystyle{named}
\bibliography{ijcai}

\end{document}